\documentclass{ecai}  

\usepackage{graphicx}
\usepackage{latexsym}
\usepackage{times}
\usepackage{soul}
\usepackage{url}
\usepackage[hidelinks]{hyperref}
\usepackage[utf8]{inputenc}
\usepackage[small]{caption}
\usepackage{amsmath}
\usepackage{amsfonts}
\usepackage{subfig}
\usepackage{booktabs}
\usepackage{algorithm}
\usepackage{algorithmic}
\usepackage[switch]{lineno}
\usepackage[justification=centering]{caption}
\usepackage{float}

\begin{document}

\begin{frontmatter}

\title{CAFIN: Centrality Aware Fairness inducing IN-processing for Unsupervised Representation Learning on Graphs}


\author[A]{\fnms{Arvindh}~\snm{Arun}\thanks{Email: arvindh.a@research.iiit.ac.in}}
\author[A]{\fnms{Aakash}~\snm{Aanegola}}
\author[A]{\fnms{Amul}~\snm{Agrawal}}
\author[B]{\fnms{Ramasuri}~\snm{Narayanam}\thanks{Email: rnarayanam@adobe.com}}
\author[A]{\fnms{Ponnurangam}~\snm{Kumaraguru}}


\address[A]{IIIT, Hyderabad}
\address[B]{Adobe Research, Bangalore, India}

\begin{abstract}
Unsupervised Representation Learning on graphs is gaining traction due to the increasing abundance of unlabelled network data and the compactness, richness, and usefulness of the representations generated. In this context, the need to consider \textit{fairness and bias} constraints while generating the representations has been well-motivated and studied to some extent in prior works. One major limitation of most of the prior works in this setting is that they do not aim to address the bias generated due to connectivity patterns in the graphs, such as varied node centrality, which leads to a disproportionate performance across nodes. In our work, we aim to address this issue of mitigating bias due to inherent graph structure in an unsupervised setting. To this end, we propose CAFIN, a centrality-aware fairness-inducing framework that leverages the structural information of graphs to tune the representations generated by existing frameworks. We deploy it on GraphSAGE (a popular framework in this domain) and showcase its efficacy on two downstream tasks - Node Classification and Link Prediction. Empirically, CAFIN consistently reduces the performance disparity across popular datasets (varying from 18 to 80\% reduction in performance disparity) from various domains while incurring only a minimal cost of fairness.
\end{abstract}

\end{frontmatter}

\section{Introduction}
\label{sec:intro}
Due to the prevalence and popularity of online social networks, network data has grown significantly, both in quantity and quality, over the years \cite{leskovec2020mining}. Such rich data can be exploited to gather information at both the individual and community levels. The influx of data having inter-personal connections (represented as graphs) has served as motivation to develop several unsupervised learning algorithms for various tasks on graphs \cite{gsage,deepgraphinfomax}. These methods leverage node features along with neighborhood information to learn node representations that do not depend on the domain of the underlying graph or the desired task at hand.

It is essential that these node representations are generated with appropriate fairness measures, especially in the context of real-world deployments, to minimize bias induced by these graph learning frameworks on downstream tasks. Accordingly, {\em fairness} in the context of trained decision-making systems has increased in popularity recently due to the numerous social distresses caused when systems not incorporating adequate fairness measures were deployed in the wild \cite{fairness-reivew1,fairness-survey}. The job platform XING is an extreme example that exhibited gender-based discrimination \cite{choudhary2022survey}.

\begin{figure}[!hbtp]
\includegraphics[width=0.99\columnwidth]{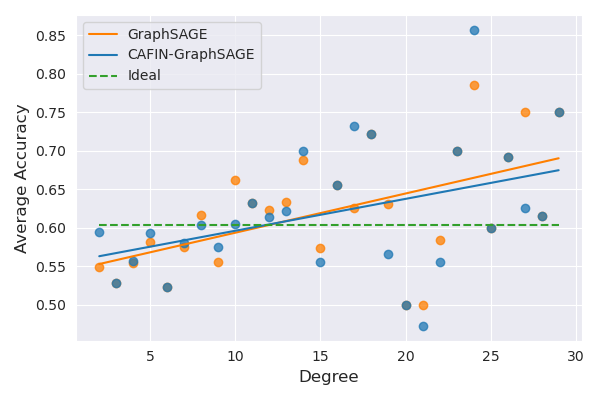}
\captionsetup{justification=centering}
\caption{\textbf{Degree vs. Accuracy plot (Twitch dataset)} - For the original GraphSAGE \cite{gsage} on the Node Classification task, the accuracy increases steadily with the degree (slope=0.0051). After the introduction of CAFIN, the slope decreases significantly (slope=0.0041, a 20\% reduction), leading to lower performance disparity between high and low degree nodes with negligible reduction (-0.3\%) in the overall accuracy.}
\label{fig:motivation}
\end{figure}

Previous works aimed to mitigate such unfairness, in this context, focus on ensuring minimal disparity in performance among individuals or groups defined by some membership criteria. Although sensitive node attributes generally decide these group memberships, a recent uptick in research considers intrinsic node properties, specifically node degree, to evaluate the fairness of Graph Neural Networks (GNNs). For example, recent work \cite{degree_gcn} provides theoretical proof that a popular subclass of graph neural networks $-$ graph convolutional networks (GCNs) $-$ are biased (in performance) towards high-degree nodes. They propose a degree-specific GCN layer targetting degree unfairness in the model and design a self-supervised learning algorithm for attaching pseudo labels to unlabelled nodes, which further helps low-degree nodes to perform better. Later, RawlsGCN \cite{rawlsgcn} reveals the root cause of this degree-related unfairness by analyzing the gradients of weight matrices in GCN and proposes techniques to mitigate this bias.

GNNs refine node embeddings by aggregating information from their neighbors. So, the efficacy of a node's representation is bound to be correlated to its abundance of structural information \cite{tailgnn}. This correlation creates a disparity in the richness of embeddings between structurally rich nodes (highly central) and the rest (less central). Figure \ref{fig:motivation} empirically corroborates this claim. This disparity is even more concerning as the centralities (degree) of most real-world graphs follow the power-law distribution. This implies that a major fraction of nodes have low centrality scores and hence deficient representations compared to a small fraction of nodes having high centrality.

Most of the works in the literature focus on imposing fairness concerning sensitive attributes but often overlook the more inherent centrality-induced disparity. Recent works \cite{liu2023generalized} also probe into how masking the sensitive attributes may not be enough, as some of the characteristics can seep into the inherent network structure. Our work in this paper focuses exclusively on reducing the performance disparity induced among groups of nodes due to skewed centrality distributions. Towards this end, we propose a generalized (additive) modification to the loss function of well-known unsupervised GNNs to impose group fairness constraints while minimizing the cost induced by the same. To formally demonstrate our approach, we consider GraphSAGE \cite{gsage} $-$ a popular unsupervised graph learning framework and widely adopted in many domains \cite{PinSage, lo2022graphsage, liu2020graphsage} $-$ and then show how we extend its objective function with fairness constraints. GraphSAGE, as studied empirically, focuses more on less frequent higher-degree nodes than on more frequent lower-degree nodes, leading to a performance disparity between the two groups of nodes. We remedy this limitation of GraphSAGE through our work.

Note that these fairness constraints can be added to any underlying graph learning algorithm at three different stages: before learning (Pre-processing), during learning (In-processing), and after learning (Post-processing) \cite{fairness-survey}. {\em In-processing} is considered robust and generalizable and finds its application across various domains as it directly adds a secondary objective to the original \cite{inprocessing}; hence we adopt this technique in our proposed framework. 

In particular, we propose a framework, Centrality Aware Fairness inducing IN-processing (CAFIN), \footnote{Code repository - \href{https://github.com/arvindh75/CAFIN}{https://github.com/arvindh75/CAFIN}} that focuses on augmenting the unsupervised version of GraphSAGE to induce centrality based (ex: degree) group fairness as an objective while maintaining similar performance on downstream tasks. {\em To the best of our knowledge, CAFIN is the first work to deal with centrality-driven fairness for unsupervised graph learning, as all other methods work in the supervised or semi-supervised setting (and also largely do not tackle centrality-based fairness aspects)}. Thus, our primary contribution is \textit{a novel in-processing technique to achieve centrality-aware group fairness for unsupervised graph node representation learning.}

\section{Related Work}
This section briefly reviews relevant literature on (unsupervised) graph representation learning and existing fairness measures for these graph representation learning algorithms.

\textbf{Graph Representation Learning.} Unsupervised representation learning on graphs has seen a recent explosion due to the availability of unlabelled structured graph data \cite{deepgraphinfomax,gsage}. In specific, GraphSAGE \cite{gsage}, a method that samples and aggregates information from node neighbors, has found extensive applications in recommender systems \cite{PinSage}, intrusion detection systems \cite{lo2022graphsage}, traffic networks \cite{liu2020graphsage}, and more due to its versatility and applicability on large graphs.

GraphSAGE \cite{gsage} is a popular inductive representation learning framework specifically tailored for efficient performance on large networks. Instead of training feature representations for each node in the graph, it learns a set of functions that aggregate feature information from the neighborhood of a node to update the node representation, helping it learn node feature embeddings while accounting for information flow from neighbors. It also uses a constrastive-learning based unsupervised loss function for learning embeddings in a task-agnostic fashion, which removes the dependence of network parameters on downstream tasks. It functions efficiently because of the random sampling in each stage of the pipeline, drastically reducing the training time as only a subset of the node neighborhood is utilized. The downside of random sampling is that it induces stochasticity in the learned embeddings, making them highly volatile and dependent on the random seed used during training \cite{stability}.

\textbf{Fairness in Graph Learning Algorithms.} The influx of deep learning technologies into the real-world setting and them leading to possibly undesired conclusions has prompted the inquisition into the fairness of the algorithms. More specific to graphs, studies like \cite{bertrand2004emily,oreopoulos2011skilled} explore the fairness of algorithms used for recruitment, and similarly \cite{khajehnejad2020adversarial} explore the issue of the unfair impact of influential nodes on the overall graph and introduce performance disparity. 

The disparity introduced by these algorithms is quantifiable, and there are two primary methods to evaluate the fairness of a graph learning algorithm - individual and group fairness. Individual fairness seeks to attain similar treatment for similar individuals \cite{dwork2012fairness}, whereas group fairness aims to reduce the bias that algorithms tend to possess towards certain groups \cite{hardt2016equality}. Group membership is usually defined based on sensitive node attributes like gender, race, and economic background in most studies \cite{krasanakis2021applying}. However, since a lot of graph data is unlabelled or does not possess sensitive node attributes, this information may not always be available. In contrast, very few studies like \cite{avin2015homophily} divide them based on node characteristics like centrality - characteristics intrinsic to the graph and present irrespective of domain. Furthermore, \cite{karimi2018homophily} confirms that degree disparities exist in real social networks, encouraging us to alleviate disparities based on intrinsic node attributes to achieve fairness.

Previous work seeks to make graph algorithms fair by (a) preprocessing the original graph to remove potential bias, for example, FairDrop \cite{spinelli2021fairdrop} that adds and removes edges to induce fairness, thereby altering graph structure, (b) in-processing during the training phase, for example, \cite{rawlsgcn} that modifies the gradient used in the optimization, and (c) postprocessing the node embeddings to remove bias \cite{fairness-survey}.

Several metrics have been proposed to evaluate group fairness along with the proliferation of methods to augment fairness. \cite{newman2003mixing} present the Assortative Mixing Coefficient, which measures communities' dependence on protected attributes and models relations between communities where connections are considered fair when the coefficient is $0$.  The notion of average statistical imparity \cite{multifair} computes the performance differences across groups but primarily caters to the two-group setting. We extend the notion of imparity to different tasks through minor modifications and utilize it for evaluation.

Previous works utilize in-processing techniques for fairer results, like \cite{rawlsgcn} that uses the Rawlsian difference principle to mitigate unfairness across the degree of Graph Convolutional Networks (GCN) and \cite{tailgnn}, which learns robust tail node (low-degree) representations by transferring information from central nodes. We address similar concerns in an unsupervised setting through CAFIN, as most prior works focus primarily on supervised and semi-supervised variants.

\begin{figure*}[h]
\includegraphics[width=2\columnwidth]{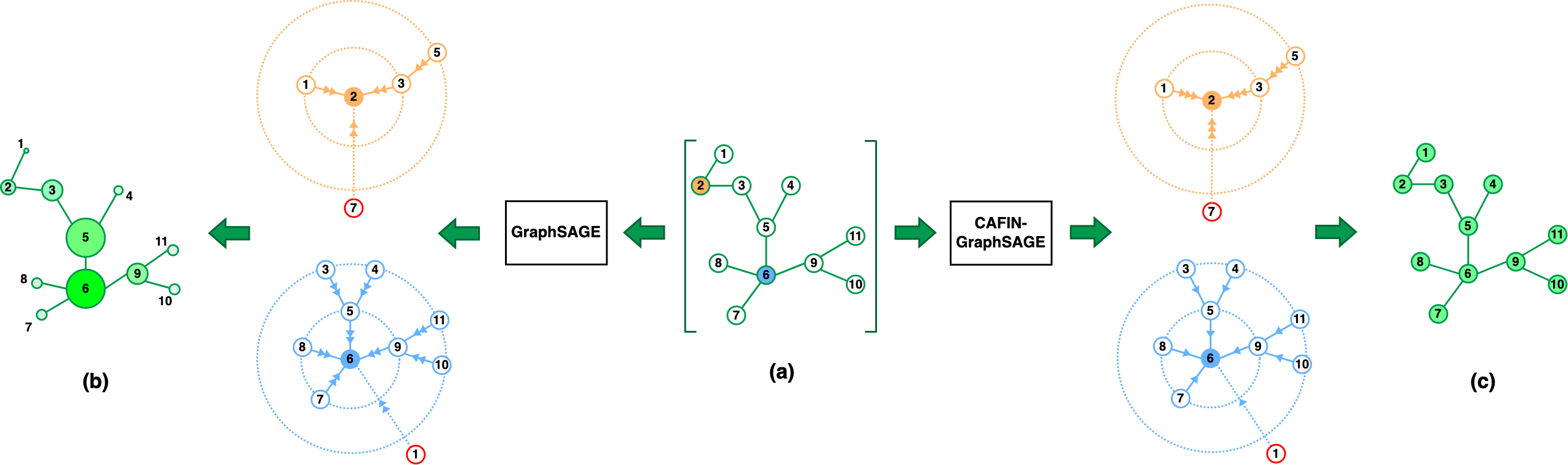}
\captionsetup{justification=centering}
\\
\caption{
\textbf{Visual Depiction of CAFIN} - GraphSAGE refines node embeddings using positive and negative samples as mentioned in section \ref{subsubsec:graphsage}. In the input graph (a), node 6 is a high-degree node (popular), while node 2 has low-degree (unpopular). As it can be noted from their computation graphs (blue and orange), node 6 has richer structural information when compared to node 2, which causes a disparity in the information flow (indicated by the arrow directions). This transitively causes a disparity in the final quality of the representations learned. The node sizes in graphs (b) and (c) represent the performance distribution in downstream tasks using the final representations learned. The introduction of CAFIN prioritizes the information flow in the computation graphs of less central nodes (indicated by stronger arrows) by penalizing them more, leading to a more homogenous distribution performance in downstream tasks, as shown in graph (c).
}
\label{fig:pipeline}
\end{figure*}

\section{Proposed Approach}
This section covers important concepts contextualizing our work and proposes our new centrality-driven fairness framework for unsupervised graph representation learning.
\label{sec:methods}

\subsection{Preliminaries}
\label{subsec:prelims}
We provide a brief overview of unsupervised representation learning, focusing on GraphSAGE, followed by group fairness (which we work with in this paper) and the evaluation metric for fairness.
\subsubsection{Unsupervised Representation Learning}
\label{subsubsec:unsupervised_representation_learning}
Unsupervised Representation Learning involves learning useful and rich features from unlabeled data. The learned representations compress and efficiently encode entity information (each node in the graph) which can later be used for several downstream tasks. Learning representations in an unsupervised fashion leads to task-agnostic representations that provide a general overview of a node's inherent characteristics, eliminating the need to re-train large networks to obtain node representations that may not translate well to other tasks. We focus on GraphSAGE, a popular unsupervised graph learning framework in our work, as it is empirically shown to have centrality-based biases. 

\subsubsection{GraphSAGE}
\label{subsubsec:graphsage}

GraphSAGE \cite{gsage} works by {\em sampling and aggregating} information from the neighborhood of each node. The sampling component involves randomly sampling $n$-hop neighbors whose embeddings are then aggregated to update the node's own embedding. It works in the unsupervised setting by sampling a positive (nearby nodes) and a negative sample (distant nodes) for each node in the training batch. It then  attempts: (a) to minimize the embedding distance between the node and the positive sample, and (b) to maximize the embedding distance between the node and the negative sample. The unsupervised version of GraphSAGE uses the following loss formulation,

\begin{equation}
\label{GraphSage_original_loss}
    \mathcal{J}_{\mathcal{G}}(z_u) = -\log(\sigma(z_u^T z_v)) - \mathcal{Q} \cdot \mathbb{E}_{v_n \sim P_n} \log(\sigma(-z_u^T z_{v_n}))
\end{equation}
where $v$ is a node that co-occurs near $u$ on fixed-length random walk, $\sigma$ is the sigmoid function, $P_n$ is a negative sampling distribution, and $Q$ defines the number of negative samples. $z_{u}, z_{v}\ \text{and}\ z_{v_{n}}$ correspond to the learned embeddings of the training sample, the positive sample, and the negative sample, respectively. The loss landscape is formulated in such a way that it is minimized when $z_{u}, z_{v}$ are close together and $z_{u}, z_{v_{n}}$ are distant in the embedding space. 

\subsubsection{Graph Centrality Measures}
\label{subsubsec:graph_centrality}
Centrality measures correlate with a node's influence on a graph and capture the relative importance of nodes \cite{centrality}. Among the several centrality measures that exist, we report results on degree centrality due to its popularity in current literature \cite{tailgnn,degree_gcn,rawlsgcn,gaps_info}. Another advantage is that the degree centrality for a graph can be calculated in linear time with respect to the number of edges, which is computationally inexpensive, unlike some of the other centralities.

\subsubsection{Centrality-driven Group Fairness}
\label{subsubsec:group_fairness}
Group Fairness concerns the disparity in the performance of a system for entities from different groups. Since most graph data does not possess explicit sensitive attributes, we utilize the connectivity structure of the graph using centrality measures to naturally categorize nodes into groups - the group of popular (centrality greater than the median) and unpopular (centrality less than the median) nodes. 

\subsection{Fairness in GraphSAGE}
\label{sec:motivation}

GraphSAGE aggregates information from its neighbors, does not consider any intrinsic structural attributes, and focuses primarily on node attributes. Intrinsic graph structure information is very valuable irrespective of domain \cite{tailgnn}, and we believe that the learning process can be made fairer by leveraging aspects of this information. GraphSAGE takes the maximum number of nodes to be sampled from each hop neighborhood as a hyperparameter to build the computation graph, which could result in central nodes having complete and larger computation graphs while the less central ones having less information-rich subgraphs. As the size of the computation graph determines how much information the chosen node aggregates and learns from its neighborhood, the representations of central nodes encode much more information, giving them an advantage over less central nodes. 

Previous works \cite{rawlsgcn} have theoretically and empirically proven the above claim for GCNs.

\begin{theorem}
\label{thm:source_of_unfairness}
Suppose we have an input graph $\mathcal{G}=\{\mathcal{V}_{\mathcal{G}}, \mathbf{A}, \mathbf{X}\}$, the renormalized graph Laplacian $\mathbf{\hat A} = \mathbf{\tilde D}^{-\frac{1}{2}} (\mathbf{A} + \mathbf{I}) \mathbf{\tilde D}^{-\frac{1}{2}}$, a nonlinear activation function $\sigma$ and an $L$-layer GCN that minimizes a task-specific loss function $J$. For any $l$-th hidden graph convolution ($\forall l\in\{1,\ldots,L\}$) layer, the gradient of the loss function $J$ with respect to the weight parameter $\mathbf{W}^{(l)}$ is a linear combination of the influence of each node weighted by its degree in the renormalized graph Laplacian.
\begin{equation*}
    \frac{\partial J}{\partial \mathbf{W}^{(l)}} 
    = \sum_{j=1}^{n} 
        \textit{deg}_{\mathbf{\hat A}}(j) 
        \mathbf{I}_j^{\textrm{(row)}}
    = \sum_{i=1}^{n} 
        \textit{deg}_{\mathbf{\hat A}}(i) 
        \mathbf{I}_i^{\textrm{(col)}}
\end{equation*}
where $\textit{deg}_{\mathbf{\hat A}}(i)$ is the degree of node $i$ in the renormalized graph Laplacian $\mathbf{\hat{A}}$, $\mathbf{I}_j^{\textrm{(row)}} = \big(\mathbf{H}^{(l-1)}[j, :]\big)^T \mathbb{E}_{i \sim p_{\mathcal{\hat N}(j)}}\bigg[\frac{\partial J}{\partial \mathbf{E}^{(l)}[i, :]}\bigg]$ and $\mathbf{I}_i^{\textrm{(col)}} = \bigg(\mathbb{E}_{j \sim p_{\mathcal{\hat N}(i)}}\big[\mathbf{H}^{(l-1)}[j, :]\big]\bigg)^T \frac{\partial J}{\partial \mathbf{E}^{(l)}[i, :]}$ are the row-wise influence matrix of node $j$ and the column-wise influence matrix of node $i$ correspondingly. $\mathbf{H}^{(l-1)}$ is the input node embeddings of the hidden layer and $\mathbf{E}^{(l)} = \mathbf{\hat A} \mathbf{H}^{(l-1)} \mathbf{W}^{(l)}$ is the node embeddings before the nonlinear activation.
\end{theorem}

Theorem \ref{thm:source_of_unfairness}, as proved in \cite{rawlsgcn}, implies that the node degree in $\mathbf{A}$ is proportional to its importance on the gradient of the weight matrix $\frac{\partial J}{\partial \mathbf{W}^{(l)}}$, implying that GCN is biased against low-degree nodes. This remark also stands true for our case as we use GraphSAGE with a mean aggregator (GraphSAGE-Mean behaves like a GCN \cite{gsage}). As described in detail in the subsequent sections, we propose a fairer learning process, CAFIN, with higher penalties for less central nodes to tackle this issue. CAFIN helps prioritize the information flow in smaller computation graphs, as depicted in Figure \ref{fig:pipeline}, alleviating the disparity caused due to the computation graph sizes.

\subsection{Imparity}
\label{subsubsec:imparity}
We focus on inter-group fairness between groups constructed from intrinsic node characteristics rather than node features. The group membership is based on degree centrality in our results, however, CAFIN can be used with any centrality measure. From Figure \ref{fig:motivation}, 
our initial analysis shows that GraphSAGE is biased towards nodes with more neighbors to learn from, and performs better for popular nodes. We use a modified variant of inter-group imparity \cite{multifair} to measure the disparity in the performance of the embeddings between unpopular and popular nodes for different downstream tasks. Fairer representations would minimize the imparity between groups. Note that imparity is not used to train the network, but rather is exclusively an evaluation method. 

\subsubsection{Imparity for Node Classification}
\label{subsubsec:node_classification}
Node classification involves identifying labels for nodes in a graph \cite{degree_gcn,gsage}. As explained earlier, we divide the nodes into two groups ($1$ $\&$ $2$) based on their centrality. We compute the inter-group accuracy differences for all classes weighted by the class distribution,
\begin{equation}
\label{eq:imparity_nc}
I_{nc} = \sum_{c \in C} w_c |a^c_1-a^c_2| 
   \quad\mathrm{and}\quad 
w_c = \frac{f_c}{|V|}
\end{equation}
where $I_{nc}$ represents the imparity for the task of node classification, $f_c$ represents the count of nodes labeled with class $c$ in the input graph, and $a^c_i$ represents the average accuracy of nodes labeled with class $c$ in the $i^{th}$ group (either popular or unpopular nodes). We use a weighted metric (where the weights are proportional to the respective class cardinality) in place of the original \cite{multifair} to avoid skewing the metric based on classes that are less common than others. In the case of \textit{multi-label node classification} (such as in PPI dataset), we compute imparity as the difference in macro-F1 scores instead of accuracy, as it is a better representative of performance in the case of multi-label data \cite{f1}.

\subsubsection{Imparity for Link Prediction}
\label{subsubsec:link_prediction}
Link Prediction involves inferring whether an edge exists between two nodes solely from their attributes and local connectivity structure \cite{grover:2016}. Based on our prior division of nodes based on popularity, edges are divided into three groups - between two popular nodes ($p-p$), between one popular and one unpopular node ($p-up$), and between two unpopular nodes ($up-up$). We define imparity as the standard deviation between the accuracies for these three types of edges.
\begin{equation}
\label{eq:imparity_lp}
I_{lp} = \sqrt{\frac{(a_{p-p} - \mu)^2 + (a_{p-up} - \mu)^2 + (a_{up-up} - \mu)^2}{3}}
\end{equation}
\begin{equation}   
\label{eq:mu_lp}
\mu = \frac{a_{p-p} + a_{p-up} + a_{up-up}}{3}
\end{equation}
where $I_{lp}$ is the imparity for the task of link prediction and $a_{x-y}$ is the accuracy of link prediction between nodes of type $x$ and $y$. This formulation ensures that the metric is minimized when equal performance is observed across all three categories of edges. We choose standard deviation (SD) over the mean absolute deviation (MAD) to emphasize the effect of extreme outliers, better quantifying the overall fairness. 

\subsection{Preprocessing the Input Graph}
\label{sec:preprocessing}
Our proposed augmentations require pre-computed centrality measures for each node and the pairwise distances between all pairs of nodes. We pre-compute the pairwise distances using a breadth-first search (BFS) from each node while incrementally computing the degree centrality values simultaneously. Our framework utilizes the pairwise distances during training to impose the fairness constraints, while the centrality values are used later for defining group membership during evaluation.

Time complexity of this step can be broken down into three components. Let $|V|$ denote the number of nodes and $|E|$ denote the number of edges. Pairwise distance calculation uses BFS from each node and incurs $O(|V|^2 +|V||E|)$ in total but can easily be parallelized for improved performance. In section \ref{subsec:ablation}, we also explore efficient approximate distance measures as a potential replacement for this step to minimize the complexity, and we observe comparable results even with approximate distance measures. We calculate the degree centralities with one pass over all the edges, which incurs $O(|E|)$.
We do not consider the centrality computation as overheads in our work, as they are used only to evaluate performance or to divide graphs that do not contribute to training time. The primary and most significant overhead is the pairwise distance computation which we consider a cost of fairness. 

\subsection{Preparing Graph Data for Inductive Setting}
\label{sec:graph-division}
To translate transductive datasets to the inductive setting, we create disjoint subgraphs for each part of the pipeline. For both the downstream tasks (node classification and link prediction), we sample three subgraphs ($g_1$, $g_2$, and $g_3$) from the original graph: One for training GraphSAGE ($g_{1}$), one for training the downstream task classifier ($g_{2}$), and the other for evaluating the classifier's performance in the downstream task ($g_{3}$). We allocate more data for training GraphSAGE ($g_1$) than the downstream task classifier ($g_2$) as it has more parameters to learn.

{\em Node Classification:} Random vertex-induced subgraphs with 60\% of the nodes for $g_{1}$, 30\% for $g_{2}$ and the rest 10\% for $g_{3}$.

{\em Link Prediction:} The subgraph for training embeddings $g_{1}$ is constructed by sampling 60\% of the edges from the original graph. Since $g_2$ and $g_3$ deal with link prediction, they need positive samples (edges that actually exist) and negative samples (fabricated edges). We split the remaining edge set into $g_{2p}$ and $g_{3p}$ randomly (the positive edge set) and construct $g_{2n}$ and $g_{3n}$, sets of artificial edges between nodes that do not have an edge in the graph (the negative edge sets). The positive and negative edge partitions are merged to obtain the graphs $g_2$ and $g_3$, $g_2 = g_{2p} \cup g_{2n}$ and $g_3 = g_{3p} \cup g_{3n}$.

\subsection{Centrality Aware Fairness Inducing In-processing (CAFIN)}
\label{subsec:fairness_augmenting_loss}
We incorporate node degree and pairwise distance measures to augment GraphSAGE's loss formulation and achieve more equitable training between popular and unpopular nodes. Since this information is not dataset-specific, our method finds applications across domains without the inclusion of any dataset-specific overhead. Our proposed novel loss formulation is described below,

\begin{equation}
    \resizebox{.95\linewidth}{!}{$
        \displaystyle
        f_l(u, v) = \frac{\max_{z}\deg(z)}{\deg(u)} \cdot \log^2\left(\dfrac{D(z_u, z_v)}{k} \cdot \dfrac{\max_{x, y}(d(x, y))}{d(u, v)}\right) \nonumber
        $}
\end{equation}

\begin{equation}
   L_{f} = f_l(u, v) + f_l(u, v_n)
\label{eq:loss_reformulation}
\end{equation}

where $\deg(u)$ represents the degree of node $u$, $z_u$ the embedding of node $u$, $D(z_u, z_v)$ the distance between the node embeddings of nodes $u$ and $v$, and $d(u, v)$ the distance between the nodes in the graph. $x$, $y$, and $z$ represent arbitrary nodes. Using GraphSAGE's notation, we represent the node of interest with $u$, the positive sample with $v$, and the negative sample with $v_n$. The parameter $k$ normalizes the embedding distance and brings it to the same range as the normalized node distance. GraphSAGE's original loss formulation takes a contrastive form that we improve by considering the actual node distance. $L_f$, the final modified loss function, converges to $0$ when the ratio between the two distances is $1$, and our loss formulation tries to make the node embedding distance equal to the actual (normalized) distance between the nodes in the graph. Most real-world graphs are assortative in nature (similar nodes are close together) \cite{assortativity}; hence, the actual node distance is a good proxy for the embedding distance. We introduce a logarithm to curtail penalties for nodes whose embedding distances are distant from the actual node distances. Additionally, we square the overall formulation to ensure that the loss reaches a minimum when the embedding distance is equivalent to the actual node distance. The loss formulation focuses more on nodes with lower degrees, which conventionally have a lower impact on learning for GraphSAGE as they have few neighbors that they influence and are influenced by. Including the inverse of node degree helps shift focus toward less popular nodes, leading to less overall disparity during the learning process. We demonstrate that these enhancements lead to a fairer version of GraphSAGE on tasks that require node representations.

\textbf{Joint Training Strategy.}
\label{subsubsec:joint_training_strategy}
To train CAFIN, we employ a joint training strategy that uses the original loss formulation as its primary objective and the modifications as its secondary.
\begin{equation}
\label{eq:training_loss}
L = L_{o} + \alpha L_{f}
\end{equation}

Equation (\ref{eq:training_loss}) describes the joint loss function where $L_o$ is the original loss $\mathcal{J}_{\mathcal{G}}(z_u)$ described in equation (\ref{GraphSage_original_loss}), and $L_f$ is the fairness-inducing constraint described in equation (\ref{eq:loss_reformulation}). $\alpha$ is a Lagrangian multiplier (balance factor) used to control the influence of the secondary fairness-inducing objective.

\section{Experimental Results}
Here we briefly describe the datasets we work with and the evaluation criteria we utilize. We then present the experimental results along with ablation studies.
\begin{table}
\centering
\caption{\textbf{Dataset Description}}
\label{tab:data-desc}
\begin{tabular}{@{}lcccc@{}}
\toprule
\textbf{Dataset} & \textbf{Nodes} & \textbf{Edges} & \textbf{Features} & \textbf{Classes} \\ \midrule
Cora     & 2,708  & 10,556  & 1,433 & 7   \\
CiteSeer & 3,327  & 9,104   & 3,703 & 6   \\
Twitch   & 7,126  & 25,468  & 20    & 2   \\
AMZN-P   & 7,650  & 238,162 & 745   & 8   \\
AMZN-C   & 13,752 & 491,722 & 767   & 10  \\
PPI      & 56,658 & 793,617 & 50    & 121 \\ 
& (24 graphs)& & & (multilbl.) \\ \bottomrule
\end{tabular}
\end{table}
\label{sec:experiments}

\begin{table*}[!ht]
\centering
\caption{\textbf{Results of CAFIN} - (a) Link Prediction, II(CV) indicates a stable increase in fairness across datasets. (b) Node Classification, II indicates an increase in fairness, however, less consistently than link prediction (as indicated by the higher CV).}
\label{tab:res}
\begin{tabular}{@{}lccc@{\hskip 0.5in}ccc@{}}
\toprule
 \multicolumn{1}{l}{}&  \multicolumn{3}{c}{(a) Link Prediction} & \multicolumn{3}{c}{(b) Node Classification}\\ 
\midrule
\textbf{Dataset} & {\textbf{II $\uparrow$ (CV $\downarrow$)}} & {\textbf{CA $\uparrow$}} & \textbf{T $\downarrow$} & {\textbf{II $\uparrow$ (CV $\downarrow$)}} & {\textbf{CA $\uparrow$}} & \textbf{T $\downarrow$}\\
\midrule
Cora     & 20.48\% (10.22\%) & -2.75\% & 0.50 & 33.13\% (10.86\%) & 0.19\%  & 0.31\\
CiteSeer & 62.89\% (15.68\%) &  3.87\% & 0.18 & 17.71\% (11.18\%) & -1.00\% & 0.65\\
Twitch   & 38.92\% (4.10\%)  & -5.67\% & 1.73 & 80.34\% (12.06\%) & -3.28\% & 0.84\\
AMZN-P   & 24.32\% (5.26\%)  & -3.30\% & 4.80 & 32.63\% (22.12\%) & -7.70\% & 3.58\\
AMZN-C   & 53.07\% (6.71\%)  & -4.56\% & 8.06 & 79.74\% (38.30\%) & -7.53\% & 5.37\\ 
PPI      & 73.31\% (12.98\%) & -3.28\% & 3.27 & 71.58\% (3.28\%)  & -3.89\% & 3.35\\ \bottomrule
\end{tabular}
\end{table*}

\subsection{Datasets}
\label{subsec:datasets}
We evaluate CAFIN on popular datasets spanning four domains, each possessing different network characteristics. Table \ref{tab:data-desc} contains the quantitative description of each dataset. We use Cora \cite{cora} and CiteSeer \cite{citeseer} from the citation network domain, Twitch (EN) \cite{twitch} dataset to study CAFIN's efficacy on social networks, co-purchase networks - Amazon Photos (AMZN-P) and Amazon Computers (AMZN-C) \cite{amzn}, and PPI \cite{ppi} from the biological networks domain. For further details on the datasets, refer to Appendix Section 1. \footnote{Appendix link - \href{https://cdn.iiit.ac.in/cdn/precog.iiit.ac.in/pubs/CAFIN_Appendix.pdf}{\text{precog.iiit.ac.in/pubs/CAFIN\_Appendix.pdf}}}

\subsection{Evaluation Criteria}
\label{subsec:evaluation_criteria}
We evaluate the effective improvement in the model's fairness by comparing the change in imparity (refer to Section \ref{subsubsec:imparity}) with the original model. The lower the imparity value of an experiment, the fairer it is compared to the original. A decrease in the imparity value indicates the reduction of the model's performance disparity between the groups, depicted by a positive percentage in the tables. We also report the change in accuracy and the time overhead, the two primary costs of fairness for CAFIN.

\textbf{Improvement in Imparity (II).}
\label{subsubsec:imparity_improvement}
The change in the imparity value measures the effective increase in fairness induced by the new formulations compared to the original. II measures the percentage decrease in imparity compared to the original. The higher the value of II, the fairer the formulation is. II is defined as,
\begin{equation*}
    II = \dfrac{I_{o} - I}{I_{o}} \cdot 100
\end{equation*}
where $I_{o}$ corresponds to the imparity values of the original and $I$ corresponds to the current imparity value.

\textbf{Change in Accuracy (CA).}
\label{subsubsec:accuracy_change}
Imposing fairness comes with a cost, like in most cases \cite{cof1,cof3}, generally in the form of a compromise in the model's performance. CA measures the overall model's accuracy change compared to the original. In an ideal experimental setting, CA will be close to $0$. CA is defined as,
\begin{equation*}
    CA = A - {A}_{o}
\end{equation*}
where $A_{o}$ corresponds to the overall accuracy of the original and $A$ corresponds to the current overall accuracy.

\textbf{Coefficient of Variance (CV).}
\label{subsubsec:variance}
We measure the consistency of our results with the Coefficient of Variance (CV), which is defined as,
\begin{equation*}
    CV = \dfrac{\sigma}{\mu} \cdot 100
\end{equation*}
where $\mu$ corresponds the mean of observed results across runs and $\sigma$ to the standard deviation. Low values of CV indicate consistency in the results. No specific ranges are considered acceptable in general as that depends on various factors like the experimental setting and objectives. \cite{cv} proposes that a CV value $\leq 10\%$ is considered excellent and anything between $10- 20\%$ is considered good in their experimental setting.

\begin{table*}[!ht]
\centering
\caption{\textbf{Results of Ablation studies for Link Prediction} - (a) For CAFIN-N, the results are less consistent than the original. (b) For CAFIN-P, the results are less consistent than the original and, in some cases, much worse. (c) For CAFIN-AD, the approximations in pairwise node distances do not impact the II or CA by much, indicating the robustness of CAFIN and reassuring the scope of scalability.}
\label{tab:ablations}
\begin{tabular}{@{}lccc@{\hskip 0.25in}ccc@{\hskip 0.25in}ccc@{}}
\toprule
 \multicolumn{1}{l}{}&  \multicolumn{3}{c}{(a) CAFIN-N} & \multicolumn{3}{c}{(b) CAFIN-P} & \multicolumn{3}{c}{(c) CAFIN-AD}\\ 
\midrule
\textbf{Dataset} & {\textbf{II $\uparrow$ (CV $\downarrow$)}} & {\textbf{CA $\uparrow$}} & \textbf{T $\downarrow$} & {\textbf{II $\uparrow$(CV $\downarrow$)}} & {\textbf{CA $\uparrow$}} & \textbf{T $\downarrow$} & {\textbf{II $\uparrow$ (CV $\downarrow$)}} & {\textbf{CA $\uparrow$}} & \textbf{T $\downarrow$}\\
\midrule
Cora     & -12.98\% (7.17\%)&-3.56\% & INF    & -216.01\% (1.10\%) & -1.41\%  & INF   & -12.68\% (14.73\%) & -5.47\%  & INF\\
CiteSeer & 84.66\% (18.17\%)&4.03\%  & 0.13   & 42.26\% (0.00\%)   & 1.92\%   & 0.26  & 77.18\% (13.01\%)  & -0.29\%  & 0.07\\
Twitch   & 28.24\% (5.35\%) &-6.96\% & 2.34   & 48.87\% (3.52\%)   & -10.04\% & 1.36  & 14.75\% (5.70\%)   & -7.26\%  & 0.34\\
AMZN-P   & 21.24\% (2.44\%) &-3.87\% & 5.46   & 23.47\% (0.82\%)   & -6.57\%  & 5.00  & 21.09\% (5.73\%)   & -4.61\%  & 0.24\\
AMZN-C   & 46.78\% (9.20\%) &-4.23\% & 9.14   & 49.66\% (9.31\%)   & -7.03\%  & 8.65  & 43.17\% (4.46\%)   & -4.15\%  & 3.59\\ 
PPI      & 90.73\% (26.14\%)&-3.52\% & 2.65   & 76.05\% (15.35\%)  & -3.01\%  & 3.16  & 24.46\% (9.91\%)   & -2.96\%  & 0.02\\ \bottomrule
\end{tabular}
\end{table*}
\addtolength{\tabcolsep}{2pt}

\textbf{Time Overhead per Increase in Imparity (T).}
\label{subsubsec:time_overhead}
This metric measures the effective increase in time per unit increase in imparity to give an idea about the effectiveness of the formulations with respect to the time overhead. Due to the augmentations, two parts in the pipeline could potentially incur a time overhead.
\begin{itemize}
    \item \textit{Training ($t_{t}$)} - We observe empirically that the time overhead in the training loop is insignificant in most cases. The increase for all datasets is less than 1\% of the original time required to train, which is in milliseconds for 100 epochs. Nevertheless, for completeness, we add it to the final time overhead.
    \item \textit{Preprocessing ($t_{p}$)} - The majority of the time overhead is constituted by preprocessing. We observe significant differences in this step as our augmentation requires extra information about the network to impose proposed constraints, specifically pairwise distance measures, which is an expensive operation. However, we propose a solution to this overhead in the form of approximate distance measures, later discussed in \ref{subsubsec:approximate_distances}. 
\end{itemize}
Based on the above two observations, we define $T$ as,
\begin{equation*}
    T = \dfrac{t}{II}
\end{equation*}
where $t = t_{t} + t_{p}$ corresponds to CPU + I/O time (in seconds) required to precompute necessary data for the augmented formulation. We divide it by II to calculate the time spent to increase II by 1\%. As the total time depends on various factors like the load on the hardware and other factors, we report the mean across 100 runs along with the CV. INF is reported when II is negative.

\subsection{Results}
The following results were obtained by using a linear SVM classifier for node classification, logistic regression for link prediction and multiclass node classification (using a one vs. rest strategy), chosen due to their prevalence in the unsupervised learning paradigm, simplicity, and performance. The classifiers require the embeddings from GraphSAGE as input and use the train/test sets which were held out during the embedding training phase. 

Table \ref{tab:res}(a) captures our results for the link prediction task using all datasets and their respective fairness costs. We observe an improvement in imparity across the board with relatively low CV values, indicating stable improvements for this task. The drop in accuracy is reasonable across datasets and even positive in the case of Citeseer (indicating that in-processing for fairness can lead to performance enhancements in the case of extreme skews in centrality distribution). Although the time overhead is significant for larger graphs, we address it by including approximate distance measures, which results in minor reductions to both II and CA but a drastic reduction in T, which makes our method more feasible for large graphs.
The low value of T, despite the size of the graph for PPI, is due to the distribution of its nodes and edges into multiple subgraphs, reducing the time overhead, more details in Appendix Section 6. 

Table \ref{tab:res}(b) contains results for the node classification task, and we observe improvement in imparity for all datasets. We also observe that the improvements are much greater than the corresponding improvements in the link prediction task, with larger drops in accuracy. Although the improvement for the node classification task is better than that of link prediction, it is also more volatile than the improvement for link prediction. The lower variance in the link prediction task results stems from the task's simplicity when compared to node classification - binary classification compared to multiclass classification. 

\subsection{Hyperparameters}
\label{subsec:hyperparameters}
As GNNs are known to be sensitive to hyperparameters, we experiment with various combinations to obtain the best-performing values for each setting. The base configuration for training is 100 epochs, a learning rate of 0.0025, and a step learning rate scheduler. We tune the learning rate for each of the datasets that we do not detail in the interest of space. We use GraphSAGE with three layers and a hidden embedding size of 256 across runs and datasets. We experimented and empirically converged on $\alpha=0.05$. We also employ a stricter negative sampling by defining a minimum distance threshold (Appendix Section 2). All training runs were performed on an NVIDIA GeForce RTX 2080 Ti and 20 Intel Xeon E5-2640 v4 CPU cores with access to a minimum of 20GB of RAM. We also empirically show the robustness of CAFIN to various seeds in Appendix Section 5.

\subsection{Ablation Studies}
\label{subsec:ablation}
We focus primarily on the loss formulation design to test which components of CAFIN leads to improvements and plausible solutions for the high time complexity of the dataset preprocessing step. We showcase the results for Link prediction here and we report the Node classification results in Appendix Section 4. We also conduct tests to prove the statistical significance of the distribution changes in these studies, further details in Appendix Section 7.

\subsubsection{Loss Formulation Design}
\label{subsubsec:exp18-19}
CAFIN treats positive and negative samples equally, but unpopular nodes have fewer positive samples than popular nodes, and the utilization of positive and negative samples may provide an unfair learning advantage to more popular nodes. To verify this theory, we construct two loss formulations based on the original hypothesis. 
\begin{equation}
\label{eq:only_pos}
L_{p}(u, v) = f_l(u, v)
\end{equation}
\begin{equation}
\label{eq:only_neg}
L_{n}(u, v_n) = f_l(u, v_n)
\end{equation}
$L_p$ (CAFIN-P) adds an additional term only for positive samples, and $L_{n}$ (CAFIN-N) adds a term for only negative samples. The model is trained jointly, similar to Eq. \ref{eq:training_loss} with the same parameter $\alpha=0.05$. From tables \ref{tab:ablations}(a) and \ref{tab:ablations}(b), it can be observed that neither formulation performs consistently across datasets 
and either compromises on the improvement in imparity or the accuracy drop. Although we observe better performances for some datasets, CAFIN remains the preferred choice due to its stability.

\subsubsection{Approximate Distance Measures}
\label{subsubsec:approximate_distances}
CAFIN and its variants require an $O(|V|^2 +|V||E|)$ overhead to compute pairwise distances for the entire graph. This computation increases the time and space requirements during the preprocessing stage, inhibiting our in-processing technique's application to larger graphs. We demonstrate results using the landmark distance method to overcome this impediment \cite{landmarkpaper}. The landmark distance method considers several ``landmarks'' that are randomly chosen and computes the distance of every node to these landmarks during preprocessing. It utilizes the distance from landmarks and uses the triangle inequality to acquire an upper bound on the distance between any two nodes during inference. The landmark method for approximate distance measures reduces the time overhead from best case $O(|V|^2)$ to $O(|V| \cdot l)$, where $l$ is the number of landmarks chosen. We consider 100 landmarks for our experiments due to performance-complexity trade-off (More details in Appendix Section 3). The time overhead can be approximated to a linear overhead for large graphs. Table \ref{tab:ablations}(c) reports the results of CAFIN using landmark approximation (CAFIN-ApproximateDistance or CAFIN-AD) in place of the original pairwise distances. We observe a nominal drop in II compared to exact distance measures but a drastic reduction in the preprocessing time required, indicating that our method is robust to aberrations in distance measures.

\section{Limitations}
Our work fills a niche, but a crucial gap in the centrality-driven fairness paradigm, so its focus is targeted. Interpretability and explainability of graph learning algorithms also need to be explored - and this work does not seek to address these concerns. Further, CAFIN does not compare and contrast the impact of various centrality measures on the fairness constraints, as it imposes only degree-centrality-based fairness constraints.

\section{Conclusions and Future Work}
\label{sec:conclusion}
We introduce CAFIN, a fairness-inducing in-processing technique, and demonstrate its efficacy in reducing degree-based disparities in the embeddings generated by GraphSAGE. CAFIN offers an average of 49.50\% and 52.52\% improvement in Imparity for Link Prediction and Node Classification tasks, respectively, across datasets. We test CAFIN's robustness by conducting various ablation studies. We also introduce the CAFIN-AD variant, which uses approximate distances for reduced computational complexity, making it highly scalable and deployable in more extensive settings. We believe that CAFIN can be extended to any other contrastive-learning-based framework in this domain. We hope our work promotes further investigation in the domain of fairness for unsupervised GNNs, explicitly focusing on graph structure-induced biases.

\bibliography{main}
\end{document}